\pdfoutput=1

\documentclass[11pt]{article}

\usepackage[final]{acl}

\usepackage{times}
\usepackage{latexsym}

\usepackage[T1]{fontenc}

\usepackage[utf8]{inputenc}

\usepackage{microtype}

\usepackage{inconsolata}

\usepackage{graphicx}
\usepackage{amsmath}
\usepackage{amssymb}
\usepackage{tikz}
\usetikzlibrary{shapes.geometric, arrows.meta, positioning, fit, backgrounds}
\usepackage{booktabs}
\usepackage{multirow}
\usepackage{array}
\title{A Survey of Automated Presentation Coaching: Systems, Methods, and Open Challenges}

\author{
    \textbf{Wen Liang$^{1,2}$}, 
    \textbf{Li Siyan$^1$}, 
    \textbf{Zackary Rackauckas$^3$},
    Julia Hirschberg$^1$
    \\
    $^1$Columbia University, USA~~~~~~
    $^2$Red Hat, USA~~~~~~
    $^3$RoleGaku, USA
    \\
    \small{\texttt{\{wl2904, siyan.li, zcr2105\}@columbia.edu}, \small{\texttt{julia@cs.columbia.edu}}}
}

\begin{document}
\maketitle
\begin{abstract}
Automated coaching for oral presentations sits at the intersection of computer-assisted pronunciation training (CAPT), prosody modeling, and speech synthesis, yet no prior work has systematically surveyed and compared existing systems along these dimensions. This survey reviews and categorizes automated presentation coaching systems, spanning pronunciation tutors, fluency and prosody coaches, multimodal trainers, and conference Q\&A practice tools. We introduce a five-dimensional task taxonomy---covering segmental pronunciation, lexical stress, suprasegmental prosody, pacing, and content faithfulness---and explicitly map surveyed systems onto it to reveal coverage gaps. We further review the core technical methods these systems employ: TTS-based exemplar generation and diagnostic methods for pronunciation, prosody, and fluency assessment. Key open challenges include the scarcity of annotated presentation corpora, achieving accent-fair feedback across diverse L1 backgrounds, and delivering low-latency diagnostics for real-time rehearsal.
\end{abstract}

\section{Introduction}
Oral presentations in English, including technical talks, research seminars, and product demos, impose demands well beyond everyday conversation. Presenters must articulate specialized terminology accurately, manage pacing relative to slide transitions, and modulate prosody so that discourse structure is clear to listeners. For English as a Second Language (ESL) speakers, these demands are compounded by the tendency to carry over sound patterns from one's native language into English, accent-specific prosodic patterns, and limited opportunities for realistic rehearsal~\cite{munro1995foreign,derwing2005second}.

The last two decades have seen growing interest in \emph{automated coaching} for oral presentations, spurred by progress in computer-assisted pronunciation training (CAPT)~\cite{witt2000gop,cucchiarini2009asr}, neural text-to-speech (TTS)~\cite{chen2024f5tts,ren2021fastspeech2}, and multimodal analysis~\cite{baltrusaitis2019multimodal}. Systems such as Rhema~\cite{tanveer2015rhema}, Mirror Mirror~\cite{schneider2015mirror}, and recent LLM-augmented Q\&A coaches~\cite{aiba2024chatgptqa} each target different facets of presentation skill. Yet the field currently lacks a unified survey that (a) catalogs and compares these systems, (b) identifies which presentation dimensions they address, and (c) exposes the problems that remain.

This paper fills that gap. We survey automated coaching systems for oral presentations, focusing on speech-based dimensions: pronunciation, lexical stress, prosody, pacing, and content faithfulness. Our scope spans both L2-specific and general public-speaking tools, since many techniques transfer across these settings. We deliberately set aside visual and gesture coaching~\cite{schneider2015mirror,baur2015becoming} except when they are integrated with speech feedback, and we do not aim to cover the full literature on general-purpose ASR or spoken language assessment.

\noindent \textbf{Contributions.} Concretely, this survey: (1)~introduces a five-dimensional task taxonomy for presentation coaching and maps surveyed systems to reveal coverage gaps; (2)~systematically reviews and categorizes existing automated presentation coaching systems; (3)~reviews the core technical methods -- TTS-based exemplar generation and diagnostic approaches -- that underpin these systems; and (4)~discusses open challenges in corpora, accent fairness, real-time deployment, and the gap between research and industry practice.

\noindent \textbf{Literature Search Strategy.} To ensure comprehensive coverage, we conducted a systematic search across ACL Anthology, IEEE Xplore, ISCA Archive, Google Scholar, and Semantic Scholar using queries combining terms such as ``presentation coaching,'' ``pronunciation training,'' ``CAPT,'' ``prosody assessment,'' ``speech fluency,'' ``TTS coaching,'' and ``L2 speaking.'' We included peer-reviewed publications from 1997 to 2025 that directly address automated coaching or assessment of oral presentation or spoken language skills. We supplemented keyword-based search with citations from key papers~\cite{witt2000gop,golonka2014technologies,aiba2024chatgptqa}, following established survey methodology~\cite{eneye2025autograding}. Studies focused exclusively on visual or gestural coaching without a speech component were excluded, as were general-purpose ASR or TTS papers. This process yielded 15 representative systems and approximately 50 supporting references spanning CAPT, prosody modeling, TTS, and educational technology.

\noindent \textbf{Organization.} Section~\ref{sec:background} reviews foundational technologies. Section~\ref{sec:taxonomy} presents the five-dimensional task taxonomy that organizes the remainder of the paper, along with the system inputs, outputs, and operating modes that connect the taxonomy to practical coaching workflows. Section~\ref{sec:survey} surveys existing systems and maps them onto the taxonomy. Section~\ref{sec:tts} reviews the core technical methods these systems employ. Section~\ref{sec:datasets} covers datasets and evaluation. Section~\ref{sec:open} discusses remaining open problems and future directions, including the gap between research and industry deployment. Section~\ref{sec:conclusion} concludes.

\section{Background: Foundational Technologies}
\label{sec:background}

We briefly review the core technologies that underpin automated presentation coaching: pronunciation assessment, prosody analysis, shadowing pedagogy, and neural TTS. See detailed discussions of diagnostic and synthesis methods in Section~\ref{sec:tts}.

\noindent \textbf{Computer-Assisted Pronunciation Training (CAPT).}
CAPT systems localize segmental errors using Goodness of Pronunciation (GOP) scores~\cite{witt2000gop} or ASR confidence measures~\cite{cucchiarini2009asr}. Earlier approaches relied on HMM forced alignment~\cite{rabiner1989hmm,franco1999automatic}; modern systems leverage CTC~\cite{graves2006ctc,cao2024ctc} and self-supervised representations~\cite{baevski2020wav2vec2,hsu2021hubert} for alignment-free mispronunciation detection~\cite{xu2021explore,peng2021gopt}. These methods form the diagnostic backbone of most pronunciation coaching systems.

\noindent \textbf{Prosody Analysis.}
Effective presentations utilize prosody (intonation, phrasing, rhythm) to signal structure~\cite{hirschberg2004pragmatics}. Listener-impression studies~\cite{shoda2023prosodyimpressions} confirm that even small pitch and timing adjustments significantly affect perceived speaker competence. Assessment typically compares log-$F_0$ contours and duration patterns between learner and reference utterances~\cite{sakoe1978dtw,rosenberg2010autobi}.

\noindent \textbf{Shadowing Pedagogy.}
Exemplar-guided shadowing (imitation) is well established for improving L2 pronunciation, prosody, and fluency~\cite{hori2008shadowing,kadota2019shadowing,hamada2018shadowing}. When mimicking short auditory demonstrations with explicit focus on speech rate and prominence, learners experience improvements in timing, stress, and intonation~\cite{hsieh2013shadowing}. This pedagogy directly motivates the use of TTS to generate controllable, adjustable exemplars for coaching at scale.

\noindent \textbf{Neural Text-to-Speech (TTS).}
Recent non-autoregressive flow-matching TTS models (e.g., F5-TTS~\cite{chen2024f5tts}, Voicebox~\cite{le2024voicebox}, CosyVoice~2~\cite{du2024cosyvoice}) synthesize highly natural speech with real-time factors below 1. These systems offer fine-grained control over speaking rate, pause insertion, and emphasis, making them well-suited for generating coaching exemplars. Zero-shot style transfer from short enrollment clips enables personalized references~\cite{jia2018transfer,casanova2022yourtts}.

\section{Presentation Coaching Taxonomy}
\label{sec:taxonomy}

\begin{figure*}[t]
\centering
\resizebox{\linewidth}{!}{%
\begin{tikzpicture}[
  font=\normalsize,
  every node/.style={rounded corners=4pt},
  root/.style={fill=gray!20, draw=gray!60, text width=5.0cm, align=center, minimum height=0.75cm},
  dim/.style={fill=blue!12, draw=blue!40, text width=3.0cm, align=center, minimum height=0.7cm},
  leaf/.style={fill=white, draw=gray!50, text width=2.9cm, align=left, minimum height=0.65cm, font=\small},
  edge from parent/.style={draw=gray!60, -Stealth},
  level 1/.style={sibling distance=3.3cm, level distance=1.5cm},
  level 2/.style={sibling distance=0cm, level distance=1.4cm}
]
\node[root] {Presentation Coaching}
  child { node[dim] {Pronunciation\\(Segmental)}
    child { node[leaf] {Phone/word\\correctness} } }
  child { node[dim] {Lexical\\Stress}
    child { node[leaf] {Syllable\\prominence} } }
  child { node[dim] {Prosody\\(Suprasegmental)}
    child { node[leaf] {F0, rhythm,\\phrasing} } }
  child { node[dim] {Pacing}
    child { node[leaf] {WPM, pauses,\\articulation rate} } }
  child { node[dim] {Content\\Faithfulness}
    child { node[leaf] {WER, keyword\\coverage} } };
\end{tikzpicture}%
}
\caption{Five-dimensional taxonomy for automated presentation coaching. Existing systems (Table~\ref{tab:systems}) cover different subsets; lexical stress and content faithfulness are the least-addressed dimensions.}
\label{fig:taxonomy}
\end{figure*}
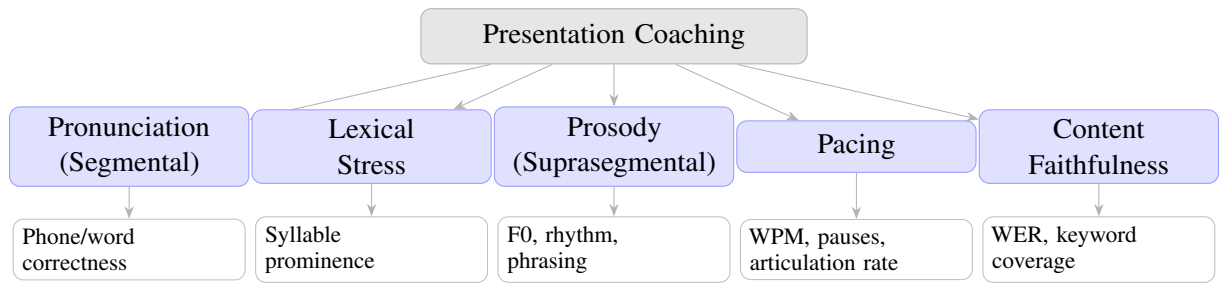

Building on the foundational technologies reviewed above, we formalize a five-dimensional taxonomy for automated presentation coaching. The taxonomy organizes presentation skills by the nature of the feedback they require and the methods available to assess them, providing a systematic framework for comparing existing systems and identifying coverage gaps. Figure~\ref{fig:taxonomy} illustrates the structure.

\subsection{Taxonomy Dimensions}

We define each dimension, its assessment methods, and the kind of feedback it yields; Section~\ref{sec:survey} maps the full set of surveyed systems onto these dimensions, and Table~\ref{tab:systems} provides the complete mapping.

\textbf{Pronunciation (segmental)} covers phone- and word-level correctness, including salient vowel/consonant contrasts and technical terminology. Assessment relies on GOP~\cite{witt2000gop}, CTC-based~\cite{cao2024ctc}, or self-supervised methods~\cite{xu2021explore}; feedback for this dimension identifies top-$k$ error phones. This is the most widely addressed dimension in the literature.

\textbf{Lexical stress} addresses syllable prominence in multi-syllabic words (e.g., \emph{AL-go-rithm}). Feedback identifies the incorrect stress position and proposes re-stress drills. Despite its documented impact on comprehensibility~\cite{munro1995foreign}, this dimension is almost entirely neglected by existing systems~\cite{korzekwa2022detection}.

\textbf{Prosody (suprasegmental)} includes intonation (F0), phrasing, rhythm, and intensity contours that signal discourse structure. Feedback often includes F0 RMSE and Pearson~$r$ vs.\ a reference. Several systems address prosody with varying granularity.

\textbf{Pacing} covers words-per-minute (WPM), articulation rate, and pause placement relative to punctuation and slide boundaries. Feedback highlights WPM deviation and pause statistics.

\textbf{Content faithfulness} measures coverage of key content without insertions or omissions, flagging missing keywords per slide. Providing the ASR with a list of expected technical terms from the slides helps it correctly transcribe domain-specific vocabulary that a general-purpose model might otherwise miss. This is the least-addressed dimension in the literature.

\subsection{Inputs, Outputs, and Assumptions}
\label{sec:io}

Beyond the five coaching dimensions, systems also differ in their input requirements and output granularity, with each input-output pattern tied to specific taxonomy dimensions. On the \textbf{input} side, systems targeting the \emph{pronunciation} and \emph{lexical stress} dimensions require, at minimum, a text transcript and a learner recording for forced alignment~\cite{witt2000gop,neri2002pedagogy,strik2009comparing,xu2021explore}. Systems addressing \emph{prosody} and \emph{pacing} additionally assume slide-aligned scripts with section boundaries to enable discourse-level analysis~\cite{hincks2005computer,chen2014automated,schneider2015mirror}. Systems targeting \emph{content faithfulness} require the reference slide content for keyword matching. Systems that support personalized exemplars across any dimension may further accept a brief enrollment clip for voice cloning~\cite{jia2018transfer,casanova2022yourtts}.

On the \textbf{output} side, feedback granularity maps directly to the taxonomy. \emph{Pronunciation} and \emph{lexical stress} systems produce per-word or per-phone error flags~\cite{witt2000gop,cao2024ctc,korzekwa2022detection}. \emph{Prosody} and \emph{pacing} systems report global scores such as F0 deviation, articulation rate, and pause frequency~\cite{shen2021fluency,saito2023comprehensibility}. \emph{Content faithfulness} systems flag missing keywords and WER per slide~\cite{aiba2024chatgptqa}. Multimodal trainers additionally output behavioral feedback on gaze and gesture~\cite{schneider2015mirror,baur2015becoming,ramanarayanan2015automatically}. Across these systems, text normalization, grapheme-to-phoneme conversion, and forced or CTC-based alignment are standard preprocessing steps, while phonetic labels from the learner are generally not required.

\subsection{System Modes}
\label{sec:modes}

Given these inputs and outputs, coaching systems operate in three complementary modes, each serving different taxonomy dimensions. \textbf{Exemplar generation} (Section~\ref{sec:tts}) produces controllable target reads per slide and for glossary items, directly supporting the \emph{pronunciation}, \emph{lexical stress}, and \emph{prosody} dimensions by providing reference audio that models correct segment production, stress patterns, and intonation contours. \textbf{Comparative assessment} (Section~\ref{sec:diagnostics}) aligns learner audio with a reference to localize deviations, enabling diagnostics across all five taxonomy dimensions---from phone-level pronunciation errors to slide-level content omissions. A \textbf{mixed-initiative practice loop} turns diagnoses into focused drills targeting the weakest dimensions (e.g., re-stress a syllable for \emph{lexical stress}; maintain a pre-equation pause for \emph{pacing}), allowing the learner to re-record and track progress. This loop is implicit in several prior systems~\cite{shen2021fluency,saito2023comprehensibility,aiba2024chatgptqa} but rarely articulated explicitly for slide-based presentations. The next section surveys existing systems and maps them onto the taxonomy dimensions defined above.

\section{Survey of Existing Presentation Coaching Systems}
\label{sec:survey}

With the taxonomy in place, we now survey automated systems that target presentation or spoken English coaching. We organize them into four categories: CAPT-based pronunciation systems, prosody and fluency coaches, multimodal trainers, and Q\&A/interaction coaches. Table~\ref{tab:systems} maps each system onto the five taxonomy dimensions.

\subsection{CAPT-Based Pronunciation Systems}

Early pronunciation coaching focused on segmental accuracy. \citet{franco1999automatic} pioneered automatic pronunciation scoring for language instruction using HMM-based forced alignment and log-likelihood scores, establishing the feasibility of system-generated feedback on individual phones. \citet{neri2002pedagogy} examined the pedagogy-technology interface in CAPT systems, showing that explicit error localization and corrective feedback improve learner outcomes beyond passive listening. \citet{strik2009comparing} compared GOP, posterior-based, and ASR-confidence approaches for automatic pronunciation error detection, finding that methods vary considerably across L1 backgrounds, motivating per-cohort calibration. More recently, \citet{xu2021explore} demonstrated that wav2vec~2.0 representations substantially outperform HMM-based systems on L2 English mispronunciation detection benchmarks. \citet{korzekwa2022detection} specifically addressed lexical stress detection in L2 English, using data augmentation and attention-based models to identify incorrect primary-stress placement in multi-syllabic words.

However, these systems share a common limitation: they operate on isolated words or short, read sentences, not on the semi-scripted, long-form speech characteristic of presentations. Feedback is typically provided post-hoc as a list of error flags, with no practice loop or exemplar generation.

\subsection{Fluency and Prosody Coaching}

Moving beyond segmental accuracy, a second line of work targets fluency and suprasegmental quality. \citet{hincks2005computer} developed one of the earliest computer-aided systems specifically targeting spoken English for academic presentations, providing visual feedback on speaking rate and pitch variation. \citet{shen2021fluency} presented an interpretable model predicting L2 fluency from acoustic features including articulation rate, pause frequency, and phonation time, identifying the features that most strongly correlate with human ratings across diverse L1 backgrounds. \citet{saito2023comprehensibility} conducted a comprehensive program on automated L2 comprehensibility assessment, arguing that comprehensibility, rather than narrow segmental accuracy, is a more robust target for coaching.

On the synthesis side, prior work~\cite{onda2024gslmaccent,onda2025prosodyAccentSim} showed that generative spoken language models can simulate and modulate foreign-accented prosody, enabling the generation of corrected exemplars that retain the learner's vocal identity while gaining similarity to target prosody patterns.
 
\citet{zechner2009automatic} developed automated scoring of non-native spontaneous speech for high-stakes testing (e-rater), covering fluency, pronunciation, and prosody holistically.

\subsection{Multimodal Presentation Trainers}

A third wave of systems extends speech coaching to the full presentation context, incorporating visual and behavioral signals together with audio. \citet{schneider2015mirror} developed \emph{Mirror Mirror}, a multimodal presentation coaching system that provides automated feedback on body language, eye contact, and filler words in addition to speech rate. \citet{baur2015becoming} presented an intelligent tutoring system that analyzes posture, gesture, and vocal quality to generate structured coaching advice for novice presenters, grounded in public speaking pedagogy~\cite{lucas2014public}. \citet{ramanarayanan2015automatically} assessed multimodal communication skills---including speech, gaze, and gesture---using behavioral signals extracted automatically, demonstrating feasibility for scalable deployment.

While the above systems combine multiple modalities, other work targets individual speech dimensions more directly. \citet{chen2014automated} evaluated presentation skills in a tutoring context using speech recognition and audience feedback. These systems demonstrate the feasibility of presentation-specific feedback loops, though they typically address only one or two of the dimensions in our defined taxonomy at a time.

\subsection{Q\&A and Interaction-Based Coaching}

The most recent systems leverage large language models to extend coaching to the interactive components of presentations. \citet{aiba2024chatgptqa} proposed a multimodal system for conference Q\&A practice, combining ASR with ChatGPT-based question generation and TTS to simulate domain-relevant follow-up questions. Learners deliver a short talk; the system transcribes it, generates content-contingent questions, and reads them aloud, enabling repeated practice of spontaneous responses. Their user study reported reducing presentation anxiety and improving perceived preparedness among first-time conference participants.

Most recently, \citet{chen2025presentcoach} introduced PresentCoach, a dual-agent system in which one agent generates benchmark presentation videos from slides using personalized voice synthesis, while another evaluates the learner's recording against these exemplars and delivers structured feedback. It is the first system to integrate
 slide-aware exemplar generation with multi-dimensional assessment in a single LLM-driven pipeline, covering four of the five taxonomy dimensions (pronunciation, prosody, pacing, and content faithfulness). Neural phonetic posteriorgrams (PPGs)~\cite{neuralppg2024} offer a complementary phone-level diagnostic suitable for such interactive loops.

\subsection{Comparison and Discussion}
\label{sec:comparison}

Together, these four system categories cover a wide arc of presentation skills, yet no single system spans all. Table~\ref{tab:systems} compares the surveyed systems along the five coaching dimensions from our taxonomy (Section~\ref{sec:taxonomy}), whether the system provides real-time feedback, and whether it targets L2 speakers specifically.

\begin{table*}[t]
\centering
\small
\renewcommand{\arraystretch}{1.25}
\begin{tabular}{@{}llccccccc@{}}
\toprule
\textbf{System} & \textbf{Pron.} & \textbf{Stress} & \textbf{Prosody} & \textbf{Pacing} & \textbf{Content} & \textbf{Real-time} & \textbf{L2-specific} \\
\midrule
\citet{franco1999automatic}       & \checkmark & & & & & & \checkmark \\
\citet{neri2002pedagogy}            & \checkmark & & & & & & \checkmark \\
\citet{hincks2005computer}             & & & \checkmark & \checkmark & & & \checkmark \\
\citet{strik2009comparing}      & \checkmark & & & & & & \checkmark \\
\citet{zechner2009automatic}   & \checkmark & & \checkmark & \checkmark & & & \checkmark \\
\citet{chen2014automated}        & \checkmark & & \checkmark & \checkmark & \checkmark & & \\
\citet{baur2015becoming}     & & & \checkmark & \checkmark & & & \\
\citet{schneider2015mirror}   & & & & \checkmark & & \checkmark & \\
\citet{ramanarayanan2015automatically}  & \checkmark & & \checkmark & & & & \\
\citet{shen2021fluency}          & \checkmark & & \checkmark & \checkmark & & & \checkmark \\
\citet{korzekwa2022detection}   & & \checkmark & & & & & \checkmark \\
\citet{xu2021explore}               & \checkmark & & & & & & \checkmark \\
\citet{saito2023comprehensibility}  & \checkmark & & \checkmark & & & & \checkmark \\
\citet{aiba2024chatgptqa}            & & & & & \checkmark & \checkmark & \checkmark \\
\citet{chen2025presentcoach}         & \checkmark & & \checkmark & \checkmark & \checkmark & & \\
\bottomrule
\end{tabular}
\caption{Comparison of existing presentation and spoken-language coaching systems across five taxonomy dimensions: ``Pron.''~=~segmental pronunciation; ``Stress''~=~lexical stress; ``Content''~=~faithfulness/content coverage; ``Real-time''~=~feedback provided during or immediately after recording. Checkmarks indicate that the system explicitly addresses the dimension.}
\label{tab:systems}
\end{table*}

Several patterns emerge from Table~\ref{tab:systems}. First, \textbf{lexical stress is almost entirely neglected}: only \citet{korzekwa2022detection} directly targets stress detection, despite its well-documented importance for L2 comprehensibility~\cite{munro1995foreign}. Second,
 \textbf{real-time feedback remains rare}: only two surveyed systems provide in-session feedback~\cite{schneider2015mirror,aiba2024chatgptqa}; most systems assess post-hoc. Third, \textbf{content faithfulness and pacing are often decoupled}: systems either check what was said (ASR-based) or how fast (WPM), but seldom do both with slide-aligned structure. Fourth, \textbf{no existing system covers all five dimensions simultaneously}: the most recent system, PresentCoach~\cite{chen2025presentcoach}, covers four (pronunciation, prosody, pacing, and content) but still omits lexical stress, illustrating that integrated, multi-dimensional coaching remains an open challenge. These gaps are compounded by limitations in available data: most CAPT corpora~\cite{zhao2018l2arctic,zhang2021speechocean762,garofolo1993timit} focus on isolated words or short sentences, lacking slide structure, domain terminology, and discourse-level features, while large-scale corpora such as LibriSpeech~\cite{panayotov2015librispeech}, Common Voice~\cite{ardila2020commonvoice}, and TED-LIUM~\cite{hernandez2018tedlium3} are not designed for L2 coaching. In summary, existing systems address individual dimensions but do not combine exemplar-based practice with fine-grained diagnostics in a unified, presentation-specific workflow~\cite{golonka2014technologies}.

\section{Coaching Systems Methods}
\label{sec:tts}
\label{sec:diagnostics}

The systems surveyed in Section~\ref{sec:survey} draw on two complementary families of methods: \textbf{TTS-based exemplar generation}, which produces reference targets for learner practice, and \textbf{diagnostic techniques}, which assess how far a learner deviates from those targets. This section reviews both families of approaches employed by prior systems.

\subsection{TTS-Based Exemplar Generation}

Several surveyed systems use synthesized or recorded model speech as a reference for learner practice~\cite{hincks2005computer,schneider2015mirror,aiba2024chatgptqa}. We review the TTS capabilities these systems exploit and the workflows they follow.

\subsubsection{TTS Capabilities for Coaching}
Modern neural TTS offers three capabilities that the surveyed systems leverage: \textbf{controllability} over rate, pauses, and emphasis; \textbf{code-switching} support for technical terms; and \textbf{zero-shot voice transfer} from short enrollment clips. Because recent flow-matching models~\cite{chen2024f5tts,le2024voicebox} and streaming architectures such as CosyVoice~2~\cite{du2024cosyvoice} achieve sub-second latency, exemplars can now be rendered on-the-fly during rehearsal. Two controls are most relevant: \emph{tempo bands} that constrain WPM, and \emph{emphasis marks} that elicit pitch accents on key terms.

\subsubsection{Workflow and Pedagogical Strategies}
Generalizing patterns from prior systems~\cite{hincks2005computer,schneider2015mirror} and grounded in shadowing pedagogy~\cite{kadota2019shadowing,hamada2018shadowing}, a typical TTS-based coaching pipeline proceeds as follows: (1)~render an \emph{anchor} exemplar at a conservative tempo (120--140 WPM) and an optional \emph{target} at a faster tempo (150--170 WPM); (2)~record the learner in short chunks (5--12 s per section); (3)~align learner audio with the reference using CTC~\cite{graves2006ctc} or DTW~\cite{sakoe1978dtw} and compute deviations; (4)~surface focused drills based on the highest-error dimensions. Across these systems, effective exemplars tend to be short ($<$12 s) to prevent cognitive overload and offer multiple speed bands and vocabulary difficulty gradations to accommodate different proficiency levels~\cite{schneider2015mirror,golonka2014technologies}. Known limitations include over-constraining style when exemplars are treated as a single correct read, and synthetic emphasis that may not match domain conventions.

\subsection{Pronunciation and Fluency Diagnostics}
While TTS provides the coaching \emph{targets}, diagnostic methods provide the \emph{assessments}~\cite{witt2012automatic,jurafsky2024speech}. The surveyed systems employ three families of diagnostic techniques, each addressing different taxonomy dimensions.

\subsubsection{GOP and CTC-Based Assessment}
Goodness of Pronunciation (GOP)~\cite{witt2000gop}, used by several surveyed systems~\cite{franco1999automatic,zechner2009automatic,strik2009comparing}, scores each phone by comparing how strongly the acoustic evidence supports the intended sound vs.\ the most likely alternative, averaged across the phone's duration. A high GOP score indicates correct pronunciation; a low score flags a likely mispronunciation. Modern ``segmentation-free'' variants~\cite{cao2024ctc} derive boundaries from CTC posteriors~\cite{graves2006ctc} rather than forced alignment, improving robustness to disfluent L2 speech. Self-supervised models (wav2vec~2.0~\cite{baevski2020wav2vec2}, HuBERT~\cite{hsu2021hubert}, WavLM~\cite{chen2022wavlm}) further advance this line~\cite{xu2021explore,peng2021gopt}, and neural PPGs~\cite{neuralppg2024} provide an alternative backbone for fine-grained phone-level scoring. Thresholds are calibrated per L1 on small rated sets, optimizing equal error rate or UAR against human annotations.

\subsubsection{Clone-and-Compare (Personalized Reference)}
To control for timbre and speaker style, several systems~\cite{onda2024gslmaccent,onda2025prosodyAccentSim} synthesize an idealized rendition in the learner's voice via voice conversion~\cite{qian2019autovc} or zero-shot TTS~\cite{jia2018transfer,casanova2022yourtts}, then compute MFCC/SSL~\cite{davis1980mfcc,baevski2020wav2vec2} distance curves aligned with monotonic DTW~\cite{sakoe1978dtw}. Distance peaks flag likely mispronunciations while controlling for speaker-specific timbre. End-to-end ASR~\cite{radford2023whisper,gulati2020conformer} also yields word durations and confidence posteriors as complementary signals.

\subsubsection{Prosody, Pacing, and Faithfulness}
Prosodic quality is computed on log-$F_0$ after voicing decisions (YAAPT~\cite{zahorian2008yaapt}): per-slide RMSE and Pearson~$r$ capture intonation match, while word- or phrase-level duration RMSE captures rhythmic alignment. These suprasegmental metrics are used by several surveyed systems~\cite{hincks2005computer,shen2021fluency,saito2023comprehensibility}. Pacing metrics (WPM deviation, articulation rate, pause precision/recall) appear in systems targeting real-time feedback~\cite{schneider2015mirror}. For content faithfulness, constrained-LM ASR with glossary bonuses reduces false omissions on domain terms~\cite{aiba2024chatgptqa}; WER and missing-keyword flags per slide are the primary metrics. The integration of these diagnostic methods into a complete coaching pipeline is detailed in Appendix~\ref{sec:appendix_system}.

\section{Datasets and Evaluation} 
\label{sec:datasets}

The diagnostic methods described above require annotated data for training and evaluation. Table~\ref{tab:corpora} summarizes major corpora and assesses their suitability for presentation coaching according to whether the following are present in the corpus: whether the learners are L2 speakers, long-form speech characteristic of presentations, accent or L1 annotations, and prosody labels (e.g. pitch accent).

\begin{table}[t]
\centering
\small
\renewcommand{\arraystretch}{1.2}
\resizebox{\columnwidth}{!}{
\begin{tabular}{@{}lrccccc@{}}
\toprule
\textbf{Corpus} & \textbf{Hrs} & \textbf{L2} & \textbf{Long} & \textbf{Accent} & \textbf{Prosody} \\
\midrule
TIMIT~\cite{garofolo1993timit}              & 5     & & & & \\
L2-ARCTIC~\cite{zhao2018l2arctic}           & 26    & \checkmark & & \checkmark & \\
Speechocean762~\cite{zhang2021speechocean762}& 70   & \checkmark & & & \checkmark \\
Speech Accent Archive~\cite{weinberger2015speech} & -- & \checkmark & & \checkmark & \\
LibriSpeech~\cite{panayotov2015librispeech} & 960   & & & & \\
Common Voice~\cite{ardila2020commonvoice}   & 1400+ & & & \checkmark & \\
TED-LIUM 3~\cite{hernandez2018tedlium3}    & 452   & & \checkmark & & \checkmark \\
GigaSpeech~\cite{chen2021gigaspeech}       & 10k   & & \checkmark & & \\
EpaDB~\cite{vidal2019epadb}               & 3     & \checkmark & & & \checkmark \\
Speak \& Improve~\cite{knill2025speakimprove} & 340 & \checkmark & & \checkmark & \\
\bottomrule
\end{tabular}
}
\caption{Major corpora used in presentation coaching research. ``Long''~=~contains long-form or discourse-level speech; ``Accent''~=~includes L1 or accent labels; ``Prosody''~=~includes prosodic annotations. No existing corpus satisfies all four criteria simultaneously.}
\label{tab:corpora}
\end{table}

As Table~\ref{tab:corpora} shows, no existing corpus simultaneously provides L2 speech, long-form presentation structure, accent labels, and prosodic annotations. TIMIT and L2-ARCTIC focus on isolated sentences; TED-LIUM 3 and GigaSpeech~\cite{chen2021gigaspeech} are long-form but not L2-specific; Common Voice provides accent labels but consists of short read sentences. Newer resources partially close the gap: EpaDB~\cite{vidal2019epadb} offers detailed phone-level pronunciation annotations for L2 Spanish speakers of English, and the recently released Speak \& Improve Corpus~\cite{knill2025speakimprove} provides approximately 340 hours of spontaneous L2 English speech with CEFR proficiency scores and diverse L1 backgrounds. However, neither includes slide structure or presentation-specific annotations. This gap directly limits the ability to train and evaluate systems on the dimensions most critical for presentation coaching, particularly discourse prosody and slide-aligned pacing. We recommend creating small, controlled \emph{presentation mini-sets}, consisting of 5--10 minute talks per speaker with slide markers, emphasis annotations, and diverse accent coverage, as a near-term community benchmark.

To evaluate coaching systems using such data, we consolidate metrics across five categories (Table~\ref{tab:metrics}), each aligned with one or more taxonomy dimensions. Importantly, the individual metrics we recommend are not novel; they are established measures drawn from prior work in prosody analysis, fluency assessment, and speech quality evaluation. Our contribution lies in their systematic alignment with the five taxonomy dimensions to form a coherent evaluation framework for presentation coaching. This alignment ensures that each taxonomy dimension has clearly defined, measurable targets, enabling researchers to evaluate coaching systems comprehensively rather than on isolated aspects.

\textbf{Segmental} metrics assess phone- and word-level pronunciation accuracy. Phone/word F1 measures the precision and recall of correctly produced segments, while unweighted average recall (UAR) ensures that performance on rare error types is not masked by frequent correct phones. Both are derived from GOP~\cite{witt2000gop} or CTC-based~\cite{cao2024ctc} diagnostic outputs.

\textbf{Prosody} metrics capture how well a learner's intonation matches the reference. Log-$F_0$ RMSE quantifies average pitch deviation, and Pearson~$r$ measures the correlation of pitch contours, reflecting whether the learner's intonation follows the same rises and falls as the target~\cite{rosenberg2010autobi,zahorian2008yaapt}.

\textbf{Pacing} metrics evaluate temporal control. WPM deviation measures how far the learner's speaking rate falls from the target band, and pause rate captures whether pauses occur at appropriate boundaries (commas, slide breaks) rather than mid-phrase~\cite{shen2021fluency,hincks2005computer}.

\textbf{Faithfulness} metrics assess content coverage. WER measures overall transcription accuracy against the script~\cite{radford2023whisper}, while glossary hit-rate tracks whether domain-critical terms are produced correctly, using constrained ASR with glossary bonuses~\cite{aiba2024chatgptqa}.

\textbf{Perceptual} metrics provide a holistic quality check. Mean Opinion Score (MOS) captures human judgments of overall speech quality, and PESQ~\cite{rix2001perceptual} offers an automated proxy. Neural MOS predictors~\cite{lo2019mosnet} can further scale these assessments without requiring human listeners for every evaluation.

\begin{table}[t]
\centering
\renewcommand{\arraystretch}{1.2}
\resizebox{\columnwidth}{!}{
\begin{tabular}{l l l}
\toprule
\textbf{Category} & \textbf{Key Metrics} & \textbf{Diagnostic Method} \\
\midrule
Segmental & Phone/Word F1, UAR & GOP~\cite{witt2000gop}, CTC \\
Prosody & Log-$F_0$ RMSE, Pearson $r$ & Pitch track + DTW \\
Pacing & WPM deviation, Pause rate & Alignment boundaries \\
Faithfulness & WER, Glossary hit-rate & Constrained ASR \\
Perceptual & MOS, PESQ & Human / Neural pred. \\
\bottomrule
\end{tabular}
}

\caption{Evaluation metrics for presentation coaching, organized by taxonomy dimension. We adopt individual metrics from prior work; we consolidate them into a unified framework aligned with the five coaching dimensions.}
\label{tab:metrics}
\end{table}

Metrics alone are insufficient without evidence that they translate into actual learner improvement, and its underlying diagnostics have to remain reliable in realistic conditions. We therefore recommend three complementary validation protocols. First, \textbf{shadowing efficacy} should be tested via randomized crossover studies comparing TTS-guided~\cite{chen2024f5tts} and unguided rehearsal, measuring whether exemplar-based practice yields measurable gains on the five taxonomy dimensions. Second, \textbf{diagnostic precision} should be evaluated by comparing GOP~\cite{witt2000gop} and clone-and-compare methods against expert annotations, since the choice of diagnostic directly affects the feedback a learner receives. Third, \textbf{robustness} should be assessed by measuring metric drift under environmental noise at multiple SNR levels, because real rehearsal settings rarely match studio-quality recording conditions.

\section{Open Problems and Research Opportunities}
\label{sec:open}

The preceding sections reveal that while individual components of presentation coaching (pronunciation diagnostics, prosody modeling, exemplar generation) have matured considerably, integrating them into comprehensive, deployable systems remains an open challenge. Rather than restating these gaps, we identify five concrete research directions with specific actionable proposals.

\paragraph{Closing the lexical stress and discourse prosody gap.} Lexical stress remains the most under-addressed taxonomy dimension (Table~\ref{tab:systems}), yet mis-stressed words are a leading cause of reduced L2 comprehensibility~\cite{munro1995foreign}. A concrete next step is to extend self-supervised mispronunciation detectors~\cite{xu2021explore,peng2021gopt} with syllable-level prominence classifiers trained on forced-aligned lexical stress annotations. For discourse prosody, current systems assess intonation at the utterance level, but presentations require section-level prosodic planning (e.g., pitch resets at topic boundaries, rising contours for rhetorical questions)~\cite{hirschberg2004pragmatics,xu2005speech}. We propose training prosodic planning models that condition on slide structure and discourse markers, enabling feedback such as ``lower your pitch at the start of a new section.''

\paragraph{Real-time, personalized coaching.} Only two of 15 surveyed systems provide real-time feedback (Table~\ref{tab:systems}), yet immediate feedback is pedagogically more effective than post-hoc reports~\cite{golonka2014technologies}. Sub-second diagnostics during live rehearsal require neural audio codecs~\cite{defossez2022encodec} and efficient alignment algorithms (e.g., subsequence DTW or streaming CTC). Beyond latency, systems should learn user-specific pacing bands and prosodic targets from minimal enrollment data rather than enforcing fixed norms~\cite{trofimovich2006learning}. A concrete proposal is to develop few-shot personalization modules that adapt pacing and prosody targets from a 2--3 minute calibration recording, enabling coaching that improves clarity without erasing the speaker's natural style.

\paragraph{Evaluation infrastructure and benchmark creation.} The field lacks standardized corpora with slide-aligned scripts, emphasis annotations, and diverse L1 coverage (Section~\ref{sec:datasets}). We recommend a community effort to create \emph{presentation mini-sets}: 20--30 speakers from 5+ L1 backgrounds, each delivering a 5--10 minute technical talk with time-stamped slide boundaries, phone-level transcriptions, and expert prosody ratings. Such a benchmark would enable head-to-head comparison of coaching systems on all five taxonomy dimensions and support shared evaluation campaigns analogous to SUPERB~\cite{yang2021superb} for speech processing. Equally important is validating that proposed metrics predict actual learner improvement, not just reference similarity; longitudinal studies measuring pre/post coaching gains on each taxonomy dimension remain rare and are critically needed.

\paragraph{LLM-augmented coaching and multimodal integration.} The success of LLM-based coaching~\cite{aiba2024chatgptqa,chen2025presentcoach} raises broader questions about how large language models can be integrated across all five taxonomy dimensions. Concrete opportunities include: (a)~generating contextualized drill prompts (``Try saying \emph{algorithm} with stress on the first syllable'') that target specific taxonomy dimensions identified by diagnostics; (b)~evaluating content faithfulness through semantic similarity rather than word-level WER, enabling tolerance for paraphrasing while still catching substantive omissions; and (c)~adapting coaching tone to individual learner profiles and proficiency levels. Recent work shows that speech LLMs can outperform prior baselines for L2 oral proficiency assessment~\cite{ma2025speechllm}, suggesting a path toward end-to-end systems that jointly assess pronunciation, prosody, and fluency. However, grounding LLM outputs in acoustically verified evidence, rather than generating plausible-sounding but inaccurate feedback, remains a key challenge. Future systems should also incorporate visual cues (slides, gestures) via vision-language models~\cite{radford2021clip,girdhar2023imagebind} to enable truly multimodal coaching. Extending coaching beyond English to other academic presentation languages (e.g., Mandarin-accented English, multilingual code-switching) requires corpora, phoneme sets, and G2P systems that do not yet exist at scale~\cite{conneau2020xlsr}.

\paragraph{Bridging the research and industry gap.} Commercial L2 pronunciation and presentation coaching applications have grown rapidly. Platforms such as ELSA Speak and Speechling target segmental pronunciation, while Yoodli and Orai focus on pacing and filler-word reduction for general public speaking. However, a significant gap exists between research prototypes and deployed systems. Commercial platforms typically address one or two taxonomy dimensions and rarely publish their technical approaches, making systematic comparison with research systems difficult. Furthermore, industry systems face practical constraints, including device heterogeneity, diverse noise environments, and the need for sub-second feedback on mobile hardware, that are rarely addressed in academic evaluations. We recommend that: (a)~future surveys explicitly compare commercial and research systems where possible; (b)~researchers engage with deployed platforms to identify real-world failure modes that controlled experiments may miss; and (c)~standardized evaluation protocols be developed that span both research and commercial systems. The fragmentation between research (which advances methods) and industry (which addresses deployment constraints) could be reduced through shared benchmarks and open evaluation challenges.

\section{Conclusion}
\label{sec:conclusion}

We have surveyed automated presentation coaching systems, organizing them through a five-dimensional taxonomy (pronunciation, lexical stress, prosody, pacing, and content faithfulness) and mapping 15 representative systems onto this framework. Our comparison (Table~\ref{tab:systems}) reveals that lexical stress and content faithfulness are dramatically under-addressed, that real-time feedback remains rare, and that no existing system integrates all five dimensions. We reviewed the core technical methods, including TTS-based exemplar generation and diagnostic approaches such as GOP/CTC, clone-and-compare, and prosody/pacing metrics, and showed how they relate to the taxonomy. Open challenges include the scarcity of annotated presentation corpora, accent-fair evaluation across diverse L1 backgrounds, the latency constraints of real-time coaching, and the gap between research prototypes and industry deployment. We hope this survey and its associated taxonomy serve as a useful reference for researchers developing the next generation of integrated, evidence-based presentation-coaching systems.

\section*{Limitations}
This survey has several limitations that should be acknowledged. First, while we aim to cover representative work at the intersection of CAPT, TTS, and L2 speaking systems, our coverage is not exhaustive; for space reasons we emphasize studies that directly inform pronunciation, prosody, pacing, and comprehensibility for slide-based L2 English presentations. Second, while we synthesize methods from TTS and CAPT domains, we focus primarily on English presentation training; the applicability to other languages and presentation styles (e.g., storytelling, persuasive speaking) requires further investigation. Third, our taxonomy and metrics are derived from existing literature and may not capture all dimensions relevant to real-world coaching scenarios. Fourth, the datasets we curate and recommend are limited in scale; large-scale presentation corpora with diverse accents and domains would strengthen evaluation. Fifth, we discuss voice cloning for personalized references but do not provide empirical comparisons of clone-and-compare vs.\ expert references across different speaker populations. Finally, while we address ethics and fairness, the practical implementation of accent-fair thresholds and privacy-preserving systems requires more extensive validation in deployed settings.

\section*{Ethics Statement}

As automated coaching systems become more widely deployed, ethical concerns around fairness, privacy, and data governance are core design requirements. Many surveyed systems collect sensitive voice data and generate personalized feedback, making the following principles essential for trustworthy deployment.

\paragraph{Accent fairness.} Calibrate thresholds per L1 cohort to account for systematic phonetic differences~\cite{flege1995second,best2007nonnative}; report subgroup metrics (e.g., precision/recall by L1) to identify potential biases; and avoid penalizing identity-marking phonetic features (e.g., rhoticity patterns, vowel quality~\cite{wells1982accents,ladefoged2015course}) when intelligibility is unaffected~\cite{munro1995foreign,derwing2005second}. Systems should help learners improve clarity without erasing accent identity~\cite{isaacs2018cognition}.

\paragraph{Voice cloning and data governance.} Obtain explicit consent before creating voice clones, clearly explaining how embeddings will be used and stored; encrypt voice embeddings both in transit and at rest; support deletion/expiry mechanisms; and allow non-clone baselines (e.g., generic TTS or expert references) for users who opt out. For data governance, disclose TTS licenses, avoid retaining raw audio longer than needed (consider on-device processing), and clearly communicate data collection practices.

\paragraph{Prosodic perception.} Listener-impression studies~\cite{shoda2023prosodyimpressions} highlight that minor pitch/timing adjustments can change perceived competence and speaker traits. Ethical coaching systems should avoid reinforcing biased mappings between prosody, accent, and personas, presenting feedback as optional stylistic guidance rather than prescriptive judgments.

\section*{Acknowledgments}
We thank Ziwei Gong for reviewing the paper and providing valuable feedback.


\bibliography{custom}
\appendix

\section{System Architecture and UI Patterns}
\label{sec:appendix_system}

Translating the methods and metrics discussed above into a usable coaching system requires careful attention to architecture, latency, and user interface design~\cite{hincks2005computer,chen2014automated}. This section outlines practical technical considerations drawing on principles from public speaking pedagogy~\cite{lucas2014public,morreale2010communication} and computer-assisted language learning~\cite{golonka2014technologies}.

A typical coaching pipeline consists of seven stages: (1) script ingest \& glossary extraction; (2) exemplar synthesis (anchor/target bands, emphasis) using modern TTS systems~\cite{chen2024f5tts}; (3) user recording in short chunks; (4) alignment (CTC timing~\cite{graves2006ctc}; DTW~\cite{sakoe1978dtw} fallback); (5) diagnostics (segmental GOP~\cite{witt2000gop}/DTW + prosody/pacing); (6) drill generation; and (7) progress tracking with per-slide thresholds. This architecture generalizes patterns found in prior L2 speaking and presentation systems~\cite{aiba2024chatgptqa,shen2021fluency,saito2023comprehensibility}, clarifying the interaction between TTS, diagnostics, and UI components.

To support interactive rehearsal, systems must meet strict latency constraints, targeting sub-second feedback per chunk. This includes TTS rendering in under 200ms for 5--8s text (achievable with non-autoregressive models like F5-TTS~\cite{chen2024f5tts}), alignment and metric computation within 300ms, and UI updates in under 200ms. Privacy-sensitive applications should prioritize on-device inference and defer computationally intensive voice cloning to an offline enrollment step with explicit consent.

Finally, effective coaching interfaces should visualize these metrics through multiple synchronized views: (1) \textbf{waveform overlay} with log-$F_0$ contours (extracted using robust pitch trackers~\cite{zahorian2008yaapt}) for comparing learner and reference prosody; (2) \textbf{word-level heatmap} color-coded by pronunciation scores (GOP~\cite{witt2000gop} or DTW~\cite{sakoe1978dtw} distance) to identify problematic segments; (3) \textbf{twin playheads} for scrubbing corresponding moments in user and reference audio; (4) \textbf{per-slide checklist} highlighting top issues (e.g., ``stress on \emph{algorithm}''); (5) \textbf{pacing gauge} showing band compliance; and (6) \textbf{one-click drills} that auto-loop on error spans for focused practice.


\end{document}